\documentclass[10pt,twocolumn,letterpaper]{article}

\usepackage{cvpr}              % To produce the CAMERA-READY version
\usepackage{graphicx}
\usepackage{amsmath}
\usepackage{amssymb}
\usepackage{booktabs}
\usepackage{times}
\usepackage{epsfig}
\usepackage{enumitem}
\usepackage{breqn}
\usepackage{mathtools}
\usepackage{textcomp}
\usepackage{gensymb}
\usepackage{multirow}
\usepackage{mathrsfs}
\newcommand{\OURNAME}{FP-LIIF\xspace}

\usepackage[pagebackref,breaklinks,colorlinks]{hyperref}

\usepackage[capitalize]{cleveref}
\crefname{section}{Sec.}{Secs.}
\Crefname{section}{Section}{Sections}
\Crefname{table}{Table}{Tables}
\crefname{table}{Tab.}{Tabs.}

\def\cvprPaperID{7933} % *** Enter the CVPR Paper ID here
\def\confName{CVPR}
\def\confYear{2023}

\begin{document}

%%%%%%%%% TITLE - PLEASE UPDATE
\title{Parameter Efficient Local Implicit Image Function Network for Face Segmentation}

% \author[1]{Mausoom Sarkar}
% \author[2]{Nikitha SR}
% \author[1]{Mayur Hemani}
% \author[1]{Rishabh Jain}
% \author[1]{Balaji Krishnamurthy}

% \affil[1]{Media and Data Science Research Lab, Adobe}
% \affil[2]{IIT Madras}
\author{Mausoom Sarkar\\
MDSR Lab, Adobe\\
% {\tt\small msarkar at adobe dot com}
\and
Nikitha SR\thanks{Work done during internship at Adobe}\\
IIT Madras\\
% {\tt\small be18b011 at smail dot iitm dot ac dot in }
\and
Mayur Hemani\\
MDSR Lab, Adobe\\
\and
Rishabh Jain\\
MDSR Lab, Adobe\\
\and
Balaji Krishnamurthy\\
MDSR Lab, Adobe\\
}
\maketitle

%%%%%%%%% ABSTRACT
\begin{abstract}
Face parsing is defined as the per-pixel labeling of images containing human faces. The labels are defined to identify key facial regions like eyes, lips, nose, hair, etc. In this work, we make use of the structural consistency of the human face to propose a lightweight face-parsing method using a Local Implicit Function network, \OURNAME. We propose a simple architecture having a convolutional encoder and a pixel MLP decoder that uses $1/26^{th}$ number of parameters compared to the state-of-the-art models and yet matches or outperforms state-of-the-art models on multiple datasets, like CelebAMask-HQ and LaPa. We do not use any pretraining, and compared to other works, our network can also generate segmentation at different resolutions without any changes in the input resolution. This work enables the use of facial segmentation on low-compute or low-bandwidth devices because of its higher FPS and smaller model size. 
\end{abstract}

%%%%%%%%% BODY TEXT
\section{Introduction}
\label{sec:intro}

Face parsing is the task of assigning pixel-wise labels to a face image to distinguish various parts of a face, like eyes, nose, lips, ears, etc. This segregation of a face image enables many use cases, such as face image editing\cite{face-edit,NeuralFace2017,zhang2019synthesis}, face e-beautification \cite{emakeup}, face swapping \cite{faceswap, transferpotrait, nirkin2018face}, face completion \cite{li2017generative}.

Since the advent of semantic segmentation through the use of deep convolutional networks\cite{long2015fully}, a multitude of research has investigated face parsing as a segmentation problem through the use of fully convolutional networks \cite{guo2018residual,jackson2016cnn,lapa,lin2019face,rtnet,liu2015multi}. 
In order to achieve better results, some methods \cite{jackson2016cnn,liu2015multi} make use of conditional random fields (CRFs), in addition to CNNs. Other methods \cite{fprnn,lin2019face}, focus on a two-step approach that predicts bounding boxes of facial regions (nose, eyes, hair. etc.) followed by segmentation within the extracted regions. Later works like AGRNET\cite{agrnet} and EAGR\cite{eagrnet} claim that earlier approaches do not model the relationship between facial components and that a graph-based system can model these statistics, leading to more accurate segmentation. 

\begin{figure}
\begin{center}
  \includegraphics[width=\linewidth]{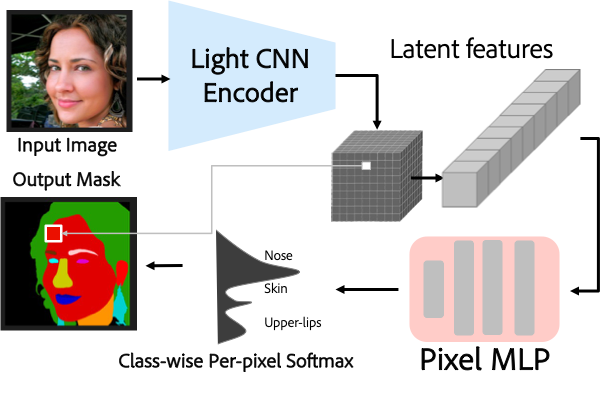}
\end{center}
\caption{The simple architecture of Local Implicit Image representation base \OURNAME: A light convolutional encoder of modified resblocks followed by a pixel only MLP decoder} 
\label{fig:banner}
\end{figure}
In more recent research, works such as FaRL\cite{farl} investigate pretraining on a human face captioning dataset. They pre-train a Vision Transformer (ViT) \cite{dosovitskiy2020image} and finetune on face parsing datasets and show improvement in comparison to pre-training with classification based pre-training like ImageNet\cite{ILSVRC15}, etc., or no pre-training at all. The current state-of-the-art model, DML\_CSR\cite{dml_csr}, tackles the face parsing task using multiple concurrent strategies including multi-task learning, graph convolutional network (GCN), and cyclic learning. The Multi-task approach handles edge discovery in addition to face segmentation. The proposed GCN is used to provide global context instead of an average pooling layer. Additionally, cyclic learning is carried out to arrive at an ensemble model and subsequently perform self-distillation using the ensemble model in order to learn in the presence of noisy labels.

\begin{figure*}[hbt!]
\begin{center}
  \includegraphics[width=\linewidth]{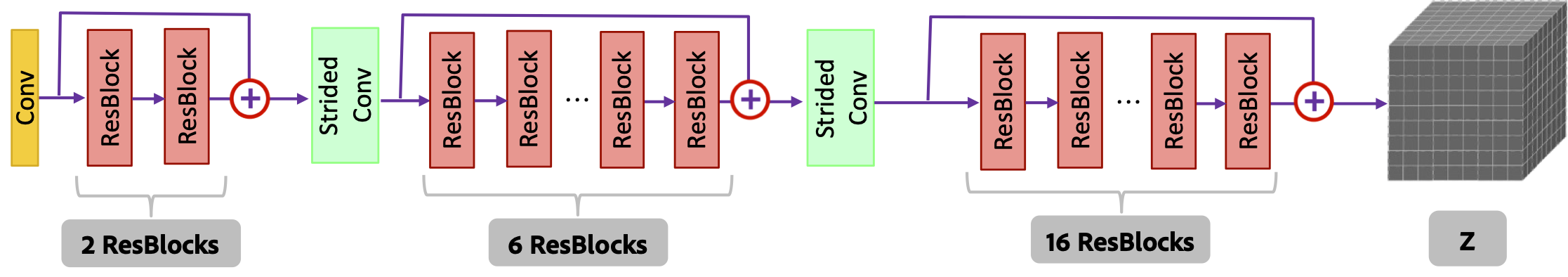}
\end{center}
\caption{Encoder Architecture: It has three res-block groups. The first two (2,6) res-block groups, followed by a strided convolution per group, are mainly used to reduce the spatial dimensions of the activation maps. The final group of res-blocks creates the grid of features vectors $Z$. Notice each res-block group has a group-level residual connection.} 
\label{fig:encoder}
\end{figure*}

\begin{figure}[!ht]
\begin{center}
  \includegraphics[width=0.9\linewidth]{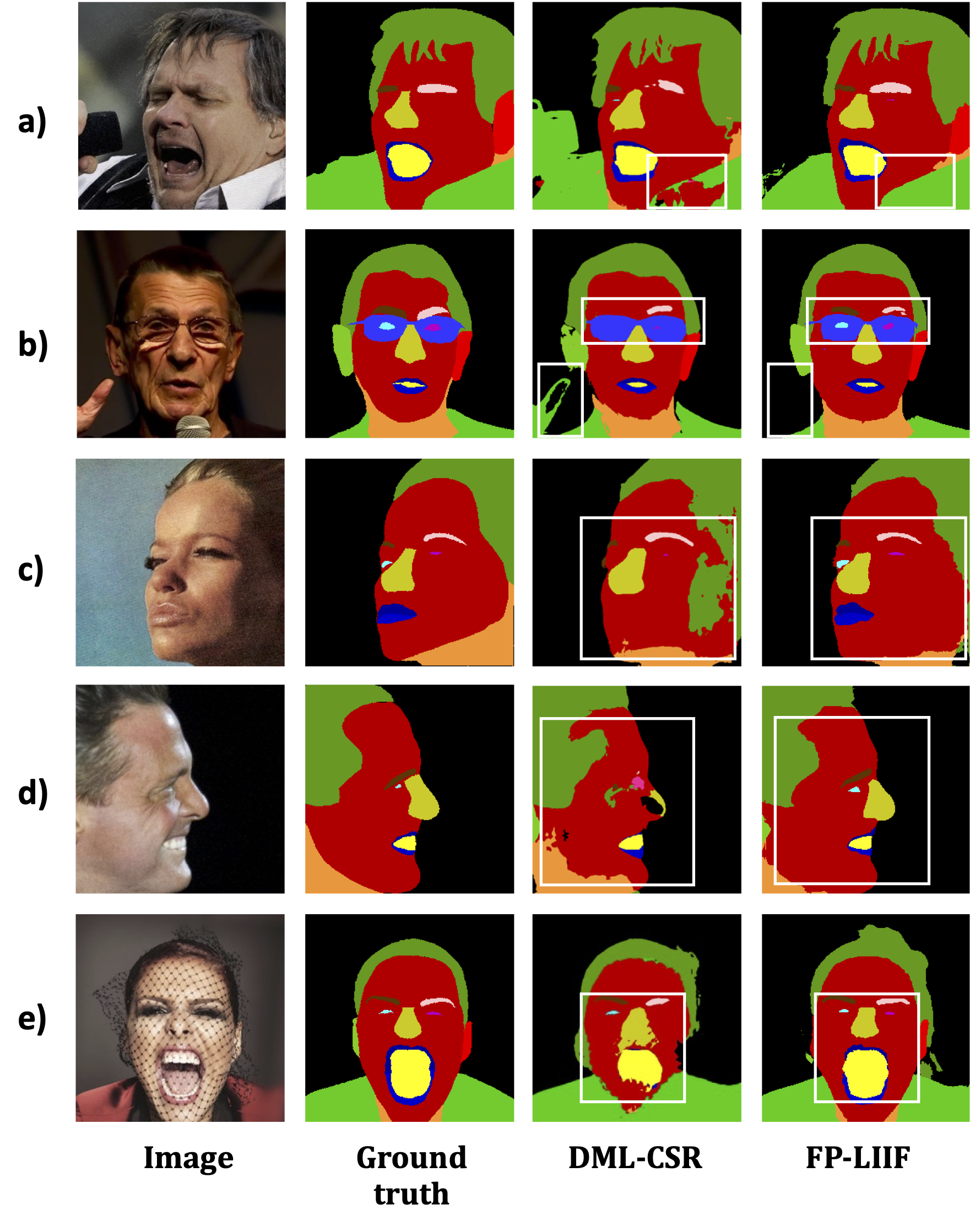}
\end{center}
\caption{Visualization of a few results in CelebAMask-HQ dataset. The difference between DML\_CSR and our results is highlighted. The cloth region in a), b), eyes in b),c) d) and nose in d),e) are better predicted by \OURNAME } 
\label{fig:celeb-we-better}
\end{figure}

\begin{figure}[!ht]
\begin{center}
  \includegraphics[width=0.9\linewidth]{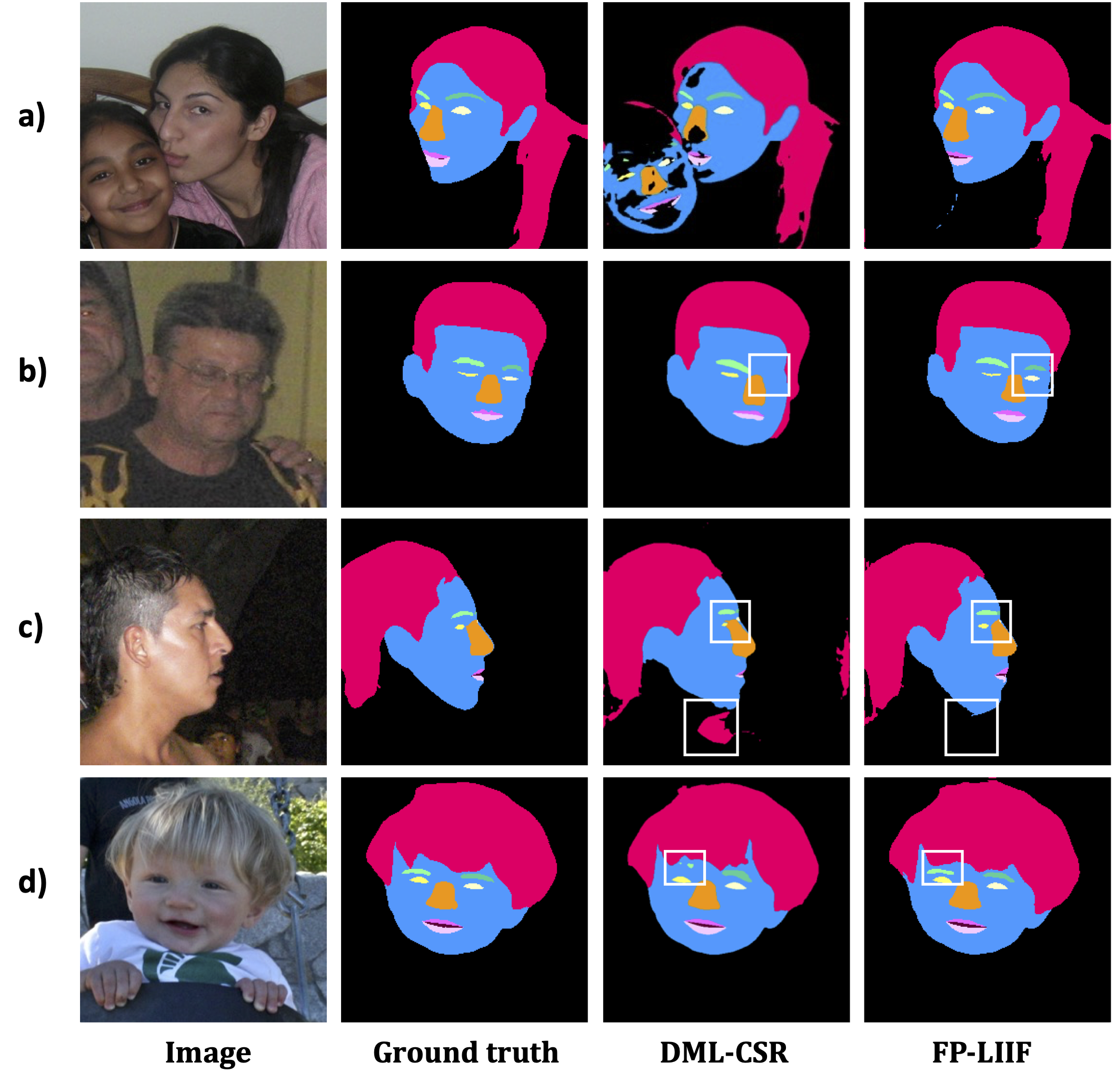}
\end{center}
\caption{Visualization of a few results in LaPa dataset. The difference between DML\_CSR and our results is highlighted. Eyes in b),c),d), Brows in b),d) are better predicted by \OURNAME } 
\label{fig:lapa-we-better}
\end{figure}

In this work, we perform face segmentation by taking advantage of the consistency seen in human facial structures. We take our inspiration from various face modeling works \cite{blanz1999morphable,gerig2018morphable,zollhofer2018state} that can reconstruct a 3D model of a face from 2D face images. These works show it is possible to create a low-dimensional parametric model of the human face in 3D. This led us to conclude that 2D modeling of the human face should also be possible with low dimension parametric model. Recent approaches, like NeRF\cite{mildenhall2020nerf} and Siren \cite{sitzmann2019siren} demonstrated that it is possible to reconstruct complex 3D and 2D scenes with implicit neural representation. Many other works \cite{ramon2021h3d,yenamandra2021i3dmm,chan2021pi,gafni2021dynamic} demonstrate that implicit neural representation can also model faces both in 3D and 2D. However, to map 3D and 2D coordinates to the RGB space, the Nerf\cite{mildenhall2020nerf} and Siren\cite{sitzmann2019siren} variants of the models require training a separate network for every scene. This is different from our needs, one of which is that we must map an RGB image into label space and require a single network for the whole domain. That brings us to another method known as LIIF\cite{liif}, which is an acronym for a Local Implicit Image Function and is used to perform image super-resolution. They learn an approximation of a continuous function that can take in any RGB image with low resolution and output RGB values at the sub-pixel level. This allows them to produce an enlarged version of the input image. Thus, given the current success of learning implicit representations and the fact that the human face could be modeled using a low-dimension parametric model, we came to the conclusion that a low parameter count LIIF-inspired model should learn a mapping from a face image to its label space or segmentation domain. In order to test this hypothesis, we modify a low-parameter version of EDSR\cite{edsr} encoder such that it can preserve details during encoding. We also modify the MLP decoder to reduce the computing cost of our decoder. Finally, we generate a probability distribution in the label space instead of RGB values. We use the traditional cross-entropy-based losses without any complicated training mechanisms or loss adaptations. An overview of the architecture is depicted in Figure \ref{fig:banner}, and more details are in Section \ref{sec:method}. Even with a parameter count that is $1/26^{th}$ compared to DML\_CSR\cite{dml_csr}, our model attains state-of-the-art F1 and IoU results for CelebAMask-HQ\cite{celebAmaskHQ} and LaPa\cite{lapa} datasets. Some visualizations of our outputs are shared in Figure \ref{fig:celeb-we-better} and Figure \ref{fig:lapa-we-better}.

To summarise, our key contributions are as follows:

\begin{itemize}[nosep]
\item We propose an implicit representation-based simple and lightweight neural architecture for human face semantic segmentation. 

\item We establish new state-of-the-art mean F1 and mean IoU scores on CelebAMask-HQ\cite{celebAmaskHQ} and LaPa\cite{lapa}.

\item Our proposed model has a parameter count of $1/26^{th}$ or lesser compared to the previous state-of-the-art model. Our model's SOTA configuration achieves an FPS of 110 compared to DML\_CSR's FPS of 76.
\end{itemize}

\section{Related Work}
\subsection{Face Parsing}
Since face parsing intrinsically involves capturing the parametric relationship between the facial regions, the existing methods in face parsing aim at modeling the spatial dependencies existing in the pixels of the image. Multiple deep learning-based models with multi-objective frameworks have been proposed to handle spatial or inter-part correlations and boundary inconsistencies and capture the image's global context. Liu et al.\cite{lapa} proposed a two-head CNN that uses an encoder-decoder framework with spatial pyramid pooling to capture global context. The other head uses the shared features from the encoder to predict a binary map of confidence that a given pixel is a boundary which is later combined with the features from the first head to perform face parsing. EAGRnet \cite{eagrnet} uses graph convolution layers to encode spatial correlations between face regions into the vertices of a graph. Zheng et al.\cite{dml_csr} combine these approaches to build DML-CSR, a dual graph convolution network that combines graph representations obtained from shallow and deeper encoder layers to capture global context. Additionally, they employ multi-task learning by adding edge and boundary detection tasks and weighted loss functions to handle these tasks. They also use an ensemble-based distillation training methodology claiming that it helps in learning in the presence of noisy labels. They achieve state-of-the-art performance on multiple face-parsing datasets. Recently a transformer-based approach has also achieved state-of-the-art performance but with the help of training with additional data. FaRL\cite{farl} starts by pre-training a model with a face-image captioning dataset to learn an encoding for face images and their corresponding captions. Their image encoder, a ViT \cite{dosovitskiy2020image}, and a transformer-based text encoder from CLIP \cite{radford2021learning} learn a common feature encoding using contrastive learning. They then use the pre-trained image encoder and finetune on various face-related task datasets to report state-of-the-art numbers. We compare our performance numbers with a non-pre-trained version of FaRL\cite{farl} because we wanted to test our model on only the task-related dataset, and using additional image-caption data was out-of-the scope of this work. \\\\
While these approaches handle the image as a whole to predict a single mask segmenting all the components simultaneously, some approaches model the individual classes separately. These approaches, called the local methods, claim that focusing on the facial components (e.g. eyes, nose, etc.) results in more accurate predictions but at the expense of efficient network structure in terms of parameter sharing. Luo et al.\cite{6247963} propose a model which segments each detected facial part hierarchically into components and pixel-wise labels. Zhou et al.\cite{Zhou_2015} built interlinking CNNs to perform localization followed by labeling. Lin et al.\cite{lin2019face} propose an RoI-Tanh operator-based CNN that efficiently uses backbone sharing and joint optimization to perform component-wise label prediction. 
Our proposed method is a whole image-based method that uses a single encoder-decoder pair to parse faces using implicit neural representations on the global image scale.

 \subsection{Parametric Human Face Models and Implicit representations}
 Parametric models for the human face have been explored for a long time since the pioneering work of \cite{parke1974parametric}. Principle component analysis (PCA) was used to model face geometry and appearance in \cite{blanz1999morphable} to create a 3D Morphable model (3DMM) of the face. Many of the approaches where the 3D face model is estimated from the 2D image, estimate coefficients of the pre-computed statistical face models\cite{thies2016face2face,ploumpis2020towards,tu2019joint}. Other methods use regression over voxels or 3D meshes\cite{feng2018joint,wei20193d} of face to arrive at a detailed face model. Many approaches have also used 3DMM type models with deep learning techniques \cite{feng2021learning,deng2019accurate,dib2021towards,lattas2020avatarme,shang2020self}.% \\\\
 
With the emergence of Implicit Neural Representation \cite{sitzmann2019siren,park2019deepsdf,mescheder2019occupancy,mildenhall2020nerf}, a new approach to parameterizing signals has been gaining popularity. Instead of encoding signals as discrete pixels, voxels, meshes, or point clouds, implicit neural representation can parameterize these as continuous functions. A lot of work in the field has been around 3D reconstruction and shape modeling \cite{chen2019learning, kellnhofer2021neural, mescheder2019occupancy,mildenhall2020nerf,park2019deepsdf,yariv2020multiview,lombardi2019neural}. Deepsdf \cite{park2019deepsdf} encodes shapes as signed distance fields and models 3D shapes with an order of magnitude lesser parameters. Neural Volume\cite{lombardi2019neural} and NeRF\cite{mildenhall2020nerf} introduced 3D volume rendering by learning a continuous function over 3D coordinates. These works led to the use of implicit representation in the domain of human body or face rendering like \cite{ramon2021h3d,yenamandra2021i3dmm}, that use implicit representation to model human heads and torso in a detailed manner. Others like Pi-Gan\cite{chan2021pi} and NerFACE\cite{gafni2021dynamic} used it in the domain of faces. NerFACE\cite{gafni2021dynamic} can extract a dynamic neural radiance field face from a monocular face video and be used with the parameters of a 3DMM. Besides 3D modeling, implicit neural representation has also been used in 2D image-to-image translation. Local Implicit Image function (LIIF) \cite{liif} proposed an implicit neural representation-based super-resolution model that treats images in the continuous domain. Based on these approaches of low-dimensional parametric face models, stunning performance of implicit neural representation in 3D reconstruction, and 2D image-to-image translation, our choice of method for exploring face segmentation gravitated towards the implicit representation approach of LIIF. The 2D texture-less appearance of the face segmentation mask prompted us to explore a low-parameter version of the LIIF model for face parsing.

\section{Methodology}
\label{sec:method}
\begin{figure}
\begin{center}
  \includegraphics[width=0.9\linewidth]{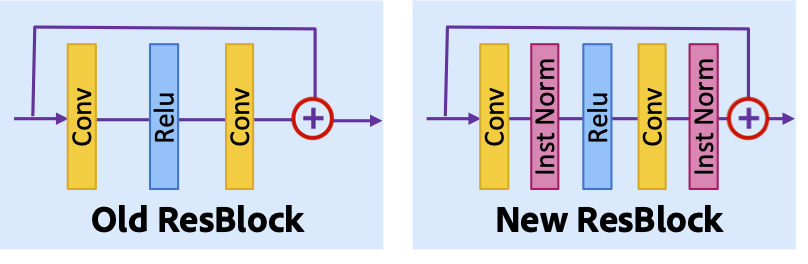}
\end{center}
\caption{ResBlock Modification: Comparison of the residual block design in EDSR with our modification. We add an Instance normalization after each convolution in the residual block.} 
\label{fig:resblock}
\end{figure}
The human face has a regular and locally consistent structure because the various features on the human face, like eyes, nose, mouth, .etc, would maintain their relative position. We use this uniformity to design a lightweight model for face parsing. We adopt the LIIF \cite{liif} framework to learn a continuous representation for segmentation for locally consistent structures of human faces.
\subsection{Segmentation as Local Implicit Image Function}
An image $I$ in LIIF is represented by a 2D grid of features $Z \in \mathbb{R}^{H\times W \times D}$ such that a function $f_{\theta}$ can map each $z\in \mathbb{R}^{D}$ in $Z$ to another domain. Here, $f$ is an MLP, and $\theta$ are its parameters. This can be represented by eq \ref{eq:liif-convert}:
\begin{equation}
\label{eq:liif-convert}
s = f_{\theta}(z, x)
\end{equation}
where, $x \in \mathcal{X}$ is a 2D coordinate, and $s$ is the signal in the domain we want to convert our image $I$ into. The coordinates are normalized in the $[-1, 1]$ range for each spatial dimension. In this paper, $s$ is the probability distribution among a set of labels, i.e., $P(y|I,x)$, where $y$ denotes the class label. So for a given image $I$, with latent codes $z$ and query coordinate $x_q$, the output can be defined as $P(y|x_q) = f_{\theta}(z,x_q)$. So using the LIIF approach, we can write
\begin{equation}
P(y|x_q) = f_{\theta}(z^*, x_q - v^*)
\label{eq:mlp}
\end{equation}
where $z^*$ is the nearest $z$ to the query coordinate $x_q$ and $v^*$ is the nearest latent vector coordinate. 
% In practise when there is a mismatch between the spatial dimensions of $Z$ (i.e. $H\times W$) and $x_q$ (i.e. $H_o\times W_o$) then $Z$ is sampled into the a grid of size $H_o\times W_o$ and used in eq. \ref{eq:mlp}.  

Other methods mentioned in LIIF\cite{liif}, such as Feature Unfolding and Local Ensemble, are also used. Feature Unfolding is a common practice of gathering local information or context by concatenating the local $z$ in the $3\times 3$ neighborhood for each $z$ in $Z$. To illustrate, the feature unfolding of a $Z$ of dimension $(H\times W \times D)$ would end up as $(H \times W \times 9D)$. Local Ensemble is a way to address the discontinuity in $f_{\theta}$ along sharp boundaries. An average of $f_{\theta}$ is calculated for each pixel according to the four nearest neighbors of $z^*$. This also bakes in a voting mechanism in the model at a per-pixel level. 

\begin{figure*}
\begin{center}
  \includegraphics[width=0.93\linewidth]{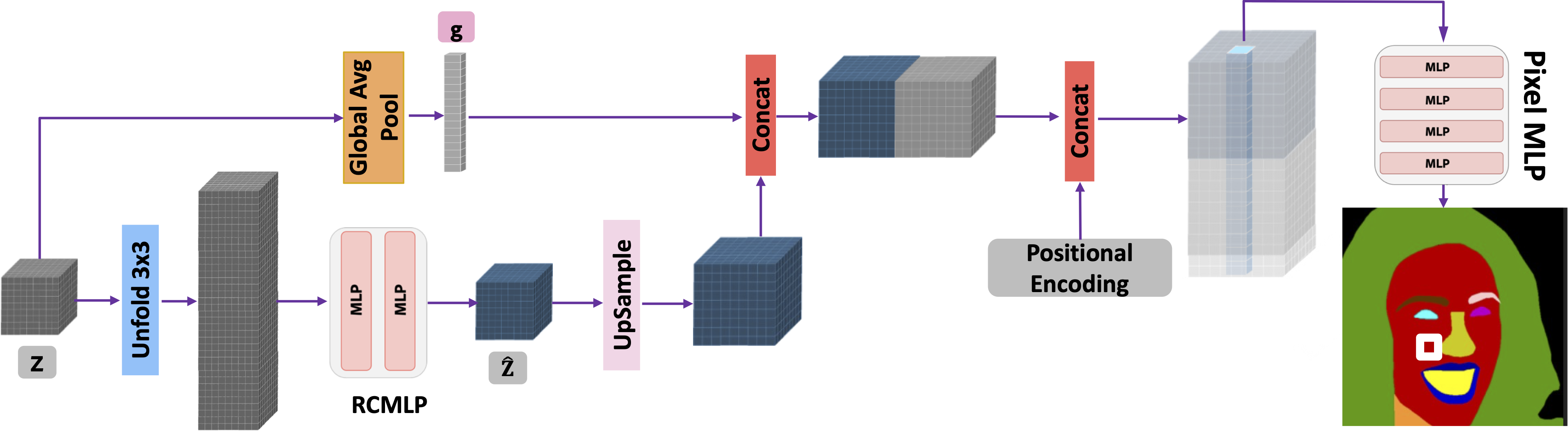}
\end{center}
\caption{Decoder Architecture: Our decoder takes in the feature grid $Z$ from the encoder and performs unfolding and global avg-pooling as shown in the figure. A two-layer fully connected MLP is used to reduce the number of channels in the unfolded volume. This is upsampled and concatenated with the global pool feature $g$ and positional encoding. Finally, a four-layer MLP is applied per spatial location to generate the classwise probability distribution.} 
\label{fig:decoder}
\end{figure*}

\subsection{Image Encoder}
We now describe our image encoder that takes as input an RGB image of size $256 \times 256$ and generates an output volume of latent vectors of size $64 \times 64$. Our encoder is a modified version of EDSR \cite{edsr} as shown in Figure \ref{fig:encoder}. We modify all the resblocks by appending an instance normalization block \cite{instnorm} after every convolution layer, Figure \ref{fig:resblock}. We create 24 resblocks Figure \ref{fig:encoder}, and all convs have a size of $3\times 3$ and filter depth of 64 channels unless otherwise stated. The input is first passed through a conv before passing it into resblock-groups. 

We have three resblock-groups. We added the first two to extract and preserve the fine-grained information from the image while the activation volume undergoes a reduction in the spatial dimensions because of the strides conv. The third group of resblock is used to generate the image representation $Z$. Each of the resblock-groups are a series of resblocks followed by a residual connection from the input, Figure \ref{fig:encoder}. The output of the first resblock-group that contains two resblocks is passed to a $3 \times 3$ conv with a stride of 2. This is passed to the second resblock-group which has six resblocks. This is again followed by a conv of stride 2. The output of this second downsampling is passed through the third resblock-group containing 16 resblocks. This generates a feature volume of size $64\times64\times64$, which is then passed to the LIIF decoder.

\begin{figure}
\begin{center}
  \includegraphics[width=0.8\linewidth]{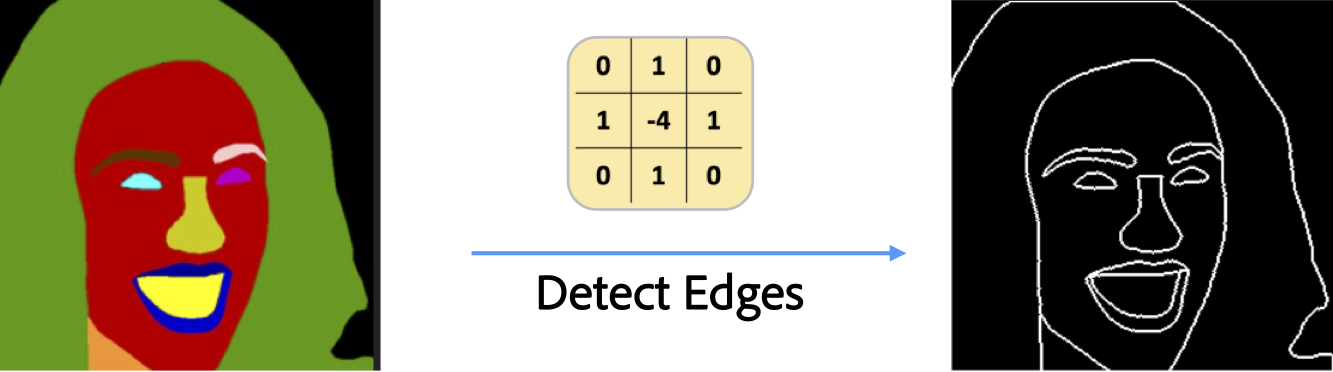}
\end{center}
\caption{Binary edge generation} 
\label{fig:edge}
\end{figure}
\subsection{LIIF decoder}
The task of the decoder is to predict the segmentation labels at each pixel, which depends on the local context and the global context. Therefore, to provide global context during per-pixel prediction, we first extract the global context by doing an average pool of the latent volume along the spatial dimensions, as shown in Figure \ref{fig:decoder}. Additional local context is added by passing the latent volume through a $3 \times 3$ unfolding operation, which increases the channel size to $64\times9$. The unfolded volume is then sent through a two-layer reduce channel MLP (RCMLP) with depths of 256 and 64. This makes the next upsampling operations computationally cheaper. 
The resulting volume $\hat{Z}$ of size $64\times64\times64$ is bilinearly upsampled to the output size  and concatenated with the previously extracted global feature and two channels of positional encoding. The positional encodings are x,y coordinates ranging from $-1$ to $1$ along the spatial dimension of the feature volume. This volume of latent vectors is flattened and passed through a 4-layer MLP of 256 channels each to predict logits for the segmentation labels.  
Next, we perform a LIIF-like ensemble of these predictions using multiple grid sampling over $\hat{Z}$. Note that the regular conv can't be used to directly replace the unfolding operation because grid sampling result would differ for a $\hat{Z}$ derived from a $w \times h \times 9D$ volume compared to a $w\times h\times D$ volume because of the different neighbors of the $D$ and $9D$ channels.

% LIIF performs an ensemble using the unfolded volume by doing a grid sampling over the $w\times h\times9D$ volume. This grid sampling result would differ for a $w \times h \times 9D$ volume and a $w\times h\times D$ volume due to differences in the neighbors of $D$ channels and unfolded $9D$ channels.

\subsection{Loss}
The logits are passed through a softmax and then guided with a cross-entropy loss $L_{cce}$ and edge-aware cross-entropy loss $L_{e\_cce}$. The edge-aware loss is calculated by extracting the edges of the ground truth label map with an edge detection kernel (Fig. \ref{fig:edge}) and calculating cross-entropy only on these edges.
The final loss can be defined as:
\begin{equation}
    L= L_{cce} + \lambda.L_{e\_cce}
\end{equation}
where $\lambda$ is the additional weight for the edge-cross-entropy.

% \begin{figure}
% \begin{center}
%   \includegraphics[width=\linewidth]{images/EdgeDetect.png}
% \end{center}
% \caption{Binary edge generation} 
% \label{fig:edge}
% \end{figure}
\section{Experiments}

\begin{table*}[!ht]
    \centering
    \scalebox{0.93}{
    \begin{tabular}{|l|l|l|l|l|l|l|l|l|l|l|l|}
    \hline
        Method/Class & Skin & Hair & Nose & I-Mouth & L-Eye & R-Eye & L-Brow & R-Brow & U-Lip & L-Lip & Mean \\ \hline
        Wei et al.\cite{afip} & 96.1 & 95.1 & 96.1 & 89.2 & 88.9 & 87.5 & 86 & 87.8 & 83.1 & 83.8 & 89.36 \\ \hline
        EAGR\cite{eagrnet} & 97.3 & 96.2 & 97.1 & 90 & 89.5 & 90 & 86.5 & 87 & 88.1 & 89 & 91.07 \\ \hline
        FARL$^{Scratch}$\cite{farl} & 97.2 & 93.1 & 97.3 & 89.4 & 91.6 & 91.5 & 90.1 & 89.7 & 87.2 & 89.1 & 91.62 \\ \hline
        AGRNET\cite{agrnet} & \textbf{97.7} & \textbf{96.5} & 97.3 & \textbf{90.7} & 91.6 & 91.1 & 89.9 & 90 & \textbf{88.5} & \textbf{90.1} & 92.34 \\ \hline
        DML\_CSR\cite{dml_csr} & 97.6 & 96.4 & \textbf{97.3} & 90.5 & 91.8 & 91.5 & 90.4 & 90.4 & 88 & 89.9 & 92.38 \\ \hline
        Ours & 97.6 & 96 & 97.2 & 90.3 & \textbf{92} & \textbf{92.2} & \textbf{90.9} & \textbf{90.6} & 87.8 & 89.5 & \textbf{92.41} \\ \hline
Ours$^{512}$ & 97.5 & 95.9 & 97.2 & 90.3 & \textbf{92} & \textbf{92.2} & \textbf{90.9} & \textbf{90.6} & 87.7 & 89.5 & \textbf{92.38} \\ \bottomrule
Ours$^{192\rightarrow256}$ & 97.5 & 96 & 97.2 & 90.3 & \textbf{92} & \textbf{92.1} & \textbf{90.8} & \textbf{90.5} & 87.7 & 89.4 & 92.35 \\ \bottomrule
Ours$^{128\rightarrow256}$ & 97.5 & 96 & 97.2 & 90.2 & {91.6} & \textbf{91.8} & \textbf{90.8} & \textbf{90.4} & 87.5 & 89.3 & {92.23} \\ \bottomrule
Ours$^{96\rightarrow256}$ & 97.4 & 95.9 & 97.2 & 90.0 & 90.1 & 90.5 & \textbf{90.5} & 90.2 & 87 & 88.9 & {91.76} \\ \bottomrule
Ours$^{64\rightarrow256}$ & 97.1 & 95.8 & 97 & 89.4 & 85.5 & 86.2 & 88.8 & 88.6 & 85 & 87.8 & {90.12} \\ \bottomrule

    \end{tabular}
    }
    \caption{Results on Lapa: F1 score comparison with baselines. Ours$^{512}$ denotes the result of generating output at 512 resolution without changing the input resolution of 256$\times$256. Ours$^{192\rightarrow256}$,Ours$^{128\rightarrow256}$,Ours$^{96\rightarrow256}$ and Ours$^{64\rightarrow256}$ denote results of upsampling from output resolution 192,128,96 and 64 respectively to a resolution of 256.}
    \label{table:lapa_results}
\end{table*}

\begin{table*}[th]
\centering
\scalebox{0.93}{
\begin{tabular}{@{}l|lllllllll|l@{}}
\toprule
\multirow{2}{*}{Method/Class} & Skin  & Nose        & E-glasses & L-Eye & R-Eye & L-Brow  & R-Brow   & L-Ear & R-Ear & \multirow{2}{*}{Mean}       \\ \cmidrule(lr){2-10}
                              & Mouth & U-Lip       & L-Lip      & Hair  & Hat   & Earring & Necklace & Neck  & Cloth &                             \\ \cmidrule(r){1-11} \cmidrule(l){11-11} 
\multirow{2}{*}{Wei et al.\cite{afip}}    & 96.4  & 91.9        & 89.5       & 87.1  & 85    & 80.8    & 82.5     & 84.1  & 83.3  & \multirow{2}{*}{82.05} \\
                              & 90.6  & 87.9        & 91         & 91.1  & 83.9  & 65.4    & 17.8     & 88.1  & 80.6  &                             \\ \cmidrule(r){1-11}
\multirow{2}{*}{FARL$^{Scratch}$\cite{farl}}         & 96.2  & 93.8        & 92.3       & 89    & 89    & 85.3    & 85.4     & 86.9  & 87.3  & \multirow{2}{*}{84.77} \\
                              & 91.7  & 88.1        & 90         & 94.9  & 82.7  & 63.1    & 33.5     & 90.8  & 85.9  &                             \\ \cmidrule(r){1-11}
\multirow{2}{*}{EAGR\cite{eagrnet}}         & 96.2  & 94          & 92.3       & 88.6  & 88.7  & \textbf{85.7}    & 85.2     & 88    & 85.7  & \multirow{2}{*}{85.14} \\
                              & \textbf{95}    & 88.9        & 91.2       & 94.9  & 87.6  & 68.3    & 27.6     & 89.4  & 85.3  &                             \\ \cmidrule(r){1-11}
\multirow{2}{*}{AGRNET\cite{agrnet}}       & 96.5  & 93.9        & 91.8       & 88.7  & 89.1  & 85.5    & 85.6     & 88.1  & \textbf{88.7}  & \multirow{2}{*}{85.53} \\
                              & 92    & 89.1        & 91.1       & 95.2  & 87.2  & 69.6    & 32.8     & 89.9  & 84.9  &                             \\ \cmidrule(r){1-11}
\multirow{2}{*}{DML\_CSR\cite{dml_csr}}     & 95.7  & 93.9        & \textbf{92.6}       & 89.4  & 89.6  & 85.5    & \textbf{85.7}     & \textbf{88.3}  & 88.2  & \multirow{2}{*}{\textbf{86.07}} \\
                              & 91.8  & 87.4        & 91         & 94.5  & \textbf{88.5}  & \textbf{71.4}    & 40.6     & 89.6  & 85.7  &                             \\ \cmidrule(r){1-11}
\multirow{2}{*}{Ours}          & \textbf{96.6}  & \textbf{94} & 92.4       & \textbf{89.6}  & \textbf{89.7}  & 85.2    & 84.9     & 86.7  & 86.6  & \multirow{2}{*}{\textbf{86.07}} \\
                              & 92.6  & \textbf{89.1}        & 91.1       & \textbf{95.2}  & 86.8  & 66.9    & \textbf{43.9}     & \textbf{91.3}  & \textbf{86.7}  &                             \\ \cmidrule(r){1-11}
\multirow{2}{*}{Ours$^{512}$} & \textbf{96.6}  & \textbf{94}    & 92.5       & \textbf{90}    & \textbf{90.1}  & 85.6    & 85.4     & 86.8  & 86.7  & \multirow{2}{*}{\textbf{86.14}} \\
                        & 92.7  & \textbf{89.4}  & \textbf{91.3 }      & \textbf{95.2}  & 86.7  & 67.2    & \textbf{42.2}     & \textbf{91.4}  & \textbf{86.8}  & \\ \bottomrule
\multirow{2}{*}{Ours$^{192\rightarrow256}$}  & \textbf{96.6}  & \textbf{94} & 92.4       & \textbf{89.6}  & \textbf{89.7}  & 85.2    & 84.9     & 86.7  & 86.6  & \multirow{2}{*}{86.05} \\
                              & 92.5  & \textbf{89.1}        & 91.1       & \textbf{95.2}  & 86.8  & 66.9    & \textbf{43.8}     & \textbf{91.3}  & \textbf{86.6}  &                             \\ \cmidrule(r){1-11}
\multirow{2}{*}{Ours$^{128\rightarrow256}$}  & \textbf{96.6}  & \textbf{93.9} & 92.4       & \textbf{89.6}  & \textbf{89.6}  & 85.2    & 84.9     & 86.7  & 86.6  & \multirow{2}{*}{86.03} \\
                              & 92.5  & 88.9        & 91       & \textbf{95.2}  & 86.8  & 66.9    & \textbf{43.9}     & \textbf{91.3}  & \textbf{86.6}  &     
                              \\ \cmidrule(r){1-11}
\multirow{2}{*}{Ours$^{96\rightarrow256}$}  & \textbf{96.5}	& \textbf{93.9}	& 92.3	& 89.3 & 89.3	& 85.1 & 84.8 & 86.7 & 86.5 &\multirow{2}{*}{85.90} \\
& 92.4	& 88.6	& 90.9	& 95.1	& 86.7	& 66.7	& \textbf{43.6}	& \textbf{91.3}  &  \textbf{86.6} & 
                            \\ \cmidrule(r){1-11}
\multirow{2}{*}{Ours$^{64\rightarrow256}$}  & \textbf{96.4}	&93.7	&92.1	&88.5	&88.5	&84.5	&84.3	&86.4	&86.3 &\multirow{2}{*}{85.52} \\
& 92.1	&87.5	&90.3	&95.1	&86.6	&65.7	&\textbf{43.7}	&\textbf{91.2}	&\textbf{86.5} &   
                              \\ \bottomrule 
\end{tabular}
}
\caption{Results on CelebAMask-HQ: F1 score comparison with baselines. Ours$^{512}$ denotes the result of generating output at 512 resolution without changing the input resolution of 256$\times$256. Ours$^{192\rightarrow256}$,Ours$^{128\rightarrow256}$,Ours$^{96\rightarrow256}$ and Ours$^{64\rightarrow256}$ denote results of upsampling from output resolution 192,128,96 and 64 respectively to a resolution of 256.}
    \label{table:celeba_results}
\end{table*}

% \begin{table}[!ht]
%     \centering
%     \begin{tabular}{|l|l|l|l|}
%     \hline
%         Dataset & CelebA MIoU $\uparrow$& LaPa MIoU $\uparrow$ \\ \hline
%         DML\_CSR & 77.81 & 87.13 \\ \hline
%         Ours$^{128}$ & \textbf{77.98} & \textbf{87.17} \\ \hline
%         Ours & \textbf{77.92} & \textbf{87.15} \\ \hline
%         Ours$^{512}$ & \textbf{78.11} & \textbf{87.13} \\ \hline
%     \end{tabular}
%     \caption{MeanIoU scores for CelebA and LaPa: Comparison of scores with DML\_CSR shows our models achieves comparable or better MIoU on both datasets while being $26$ times smaller in parameter count }
%     \label{meaniou}
% \end{table}
\begin{table}[!ht]
\centering
\scalebox{0.9}{

\begin{tabular}{|c|c c c c|c|}
\toprule
\multirow{2}{*}{Model/Class} & Skin  & Nose &U-lip & I-mouth & \multirow{2}{*}{Overall}       \\ \cmidrule(lr){2-5}
                              & L-Lip  & Eyes  & Brows   & Mouth &                             \\ \cmidrule(r){1-6} \cmidrule(l){6-6} 
\multirow{2}{*}{EAGR} & 94.6 & 96.1 & 83.6  & 89.8 & \multirow{2}{*}{93.2} \\ \cmidrule(lr){2-5}
&91    & 90.2 & 84.9  & 95.5  &     
\\ \cmidrule(lr){1-6}
\multirow{2}{*}{DML-CSR} & 96.6 & 95.5 & 87.6  & 91.2  & \multirow{2}{*}{93.8} \\ \cmidrule(lr){2-5}
& 91.2  & 90.9 & 88.5  & 95.9  &     
\\ \cmidrule(lr){1-6}
\multirow{2}{*}{Ours} & 95.1 & 94   & 79.7  & 86.3  & \multirow{2}{*}{91.2} \\ \cmidrule(lr){2-5}
& 87.6  & 89.1 & 81    & 93.6   &     
\\ \bottomrule 
% DML-CSR      & 96.6 & 95.5 & 87.6  & 91.2    & 91.2  & 90.9 & 88.5  & 95.9  & 93.8    \\ \hline
% Ours         & 95.1 & 94   & 79.7  & 86.3    & 87.6  & 89.1 & 81    & 93.6  & 91.2    \\ \hline
\end{tabular}
}
\caption{Results on Helen: F1 score comparison with baselines.Our results on non aligned Helen face data set are comparable to SOTA.} 
\label{table:helen_results}
\end{table}

\begin{table}[!ht]
    \centering
    \scalebox{0.86}{
    \begin{tabular}{|c|c|c|c|c|}
    \hline
        Model & Params $\downarrow$  & $\times$\OURNAME $\downarrow$ &GFlops $\downarrow$ &FPS$\uparrow$ \\ \hline
        DML\_CSR & $59.67$ M & $26$ & $253$ & $76$\\ \hline
        EAGR & $66.72$ M & $29$ &$235$ & $71$\\ \hline
        FARL & $150$ M & $65$ & $370$ & $26$\\ \hline
        \OURNAME(ours) & $\textbf{2.29}$ M & $\textbf{1}$ & $\textbf{85}$ & $\textbf{110}$ \\ \hline
    \end{tabular}
    }
    \caption{Model size comparison: The table shows the parameter count, GFlops, and FPS for each of the models and the relative size of each model compared to \OURNAME}
    \label{modelsize}
\end{table}
\begin{figure}
\begin{center}
  \includegraphics[width=0.9\linewidth]{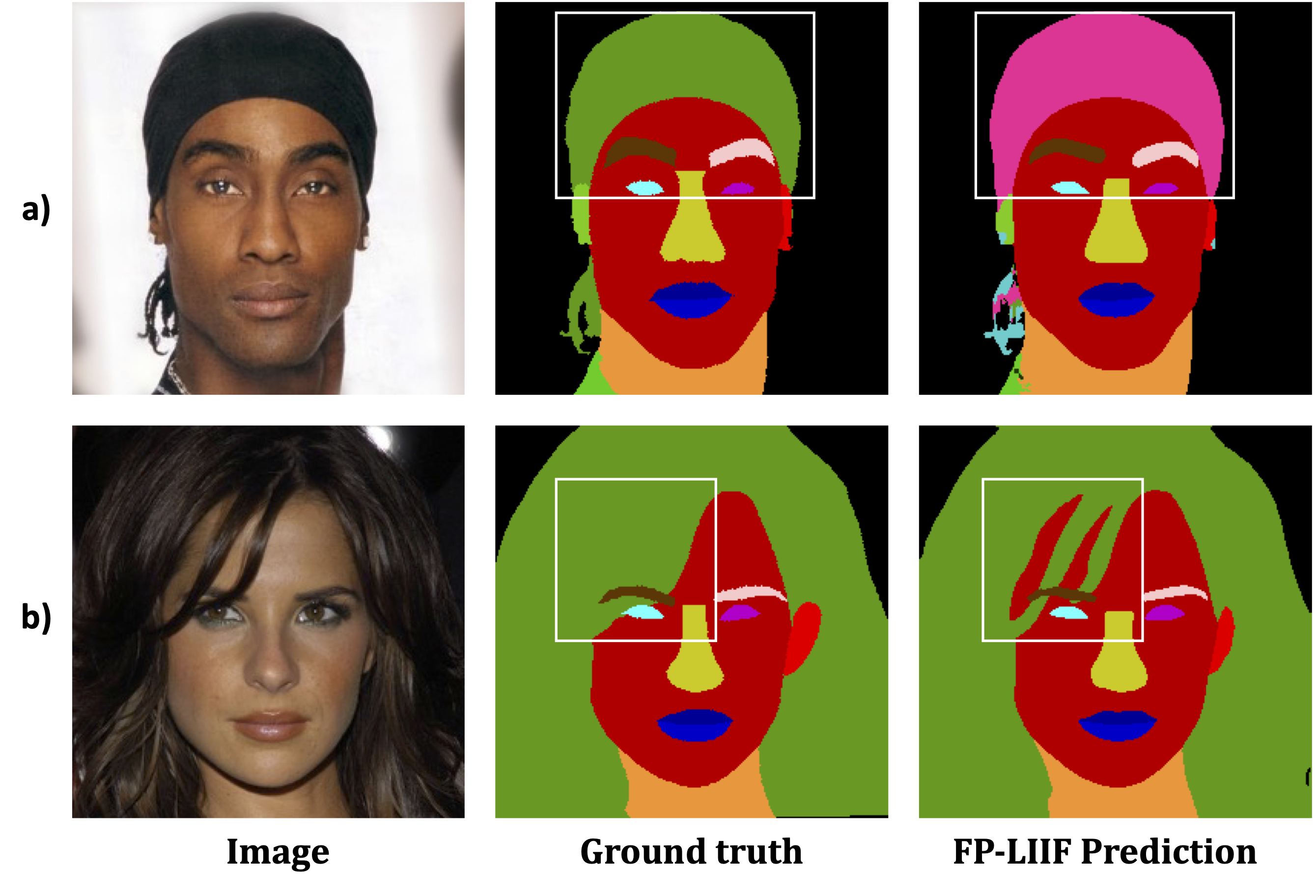}

\end{center}
\vspace{-0.5cm}
\caption{Few test samples from CelebAMask-HQ dataset illustrating noisy ground truth data and our prediction for the same. In the top row headgear has been marked as hair and in the bottom row strands of hair are not clearly segmented in the ground truth mask.} 
\vspace{-0.5cm}
\label{fig:gt-worse}

\end{figure}

\subsection{Datasets} We use three face datasets to perform our experiments, LaPa\cite{lapa}, CelebAMask-HQ \cite{celebAmaskHQ} and Helen \cite{helen}. LaPa is a face dataset with more in-the-wild photos having varying poses and occlusions. It has 22,168 images, of which 18,176 are for training, 2000 are for validation, and 2000 are for testing. The segmentation masks in LaPa have 11 categories: skin, hair, nose, left/right eyes, left/right brows, and upper/lower lips. The CelebAMask-HQ dataset contains 30k face images split into 24,183 training, 2993 validation, and 2824 test samples. It has a total of 19 semantic labels, including labels for accessories like eyeglasses, necklaces, earrings, etc., in addition to the labels in LaPa. Helen is the smallest of these three with 11 categories and has 2330 images with 2000, 230, and 100 in the training, validation, and test samples respectively. 

\subsection{Implementation Details} We implement our training and evaluation pipeline on PyTorch\cite{pytorch} version 1.10.0 with the CUDA 11.1 backbone on Python 3.8.5. We train \OURNAME on 4 Nvidia A-100 GPUs with a mini-batch size of 33 and 64 for CelebAMask-HQ and LaPa respectively. The network is optimized for 400 epochs using Adam \cite{kingma2014adam} with an initial learning rate of 5e-4. The learning rate is decreased by a factor of 10 after every 20 epochs. The $\lambda$ for edge-cross-entropy was set to 10 and 40 for CelebA and Lapa, respectively. Temperature scaling of softmax is also done with $\tau=0.5$. The images and masks in the datasets were resized to $256 \times 256$ using bicubic sampling before being used for training and evaluation. During training various data augmentations are applied like random affine transformations of rotation by 30\degree, shear between 0 to 20, scaling between 0.5 to 3 followed by random cropping. Color jitter is also applied to the input image with a brightness between [0.5,1.5], contrast [0,2], saturation [0.5,1.5], and hue [-0.3,0.3]. 

\subsection{Evaluation Metrics} 
To keep our evaluation consistent with other works, we primarily use a class-wise F1 score and a mean F1. In addition to that, we use mean intersection over union (mIoU) to compare with DML\_CSR. The background class is ignored in all these metrics.

\subsection{Baselines}
We compare our \OURNAME performance with several baselines, like Wei et al. \cite{afip} (figures taken from \cite{farl}), AGRNET\cite{agrnet}, EAGR\cite{eagrnet}, DML\_CSR\cite{dml_csr}, FARL\cite{farl} from scratch, i.e., no pre-training. The results on LaPa are reported in Table \ref{table:lapa_results}, results comparing performance on CelebAMask-HQ are in Table \ref{table:celeba_results} and results on Helen are in Table \ref{table:helen_results}. Finally, a comparison of model size, Gflops and FPS is made in Table \ref{modelsize}.
% To have a fair comparison with FARL we only compare with the $scratch$ version reported in FARL\cite{farl}, that doesn't use their image-text-caption pretraining. 

% Please add the following required packages to your document preamble:
% \usepackage{booktabs}
% \usepackage{multirow}

\section{Results} 

According to Table \ref{table:lapa_results}'s LaPa results, \OURNAME performs better overall in mean-F1 and in classes such as eyes (left-eye and right-eye), brows (left-brow and right-brow), and skin. Table \ref{table:celeba_results} demonstrates that our approach performs better than the baselines on CelebA in terms of mean-F1 and also at the class-level F1 of skin, nose, eyes (left-eye, right-eye), lips(upper-lips, lower-lips), hair, necklace, neck, and cloth. We have also included a row of results demonstrating our performance when we change our output size to $512 \times 512$. The results show that even without training for a higher resolution output, our network seamlessly generates decent segmentation results at a higher resolution with nominal degradation in LaPa while still matching the current SOTA of $92.38$ by DML\_CSR. Table \ref{table:celeba_results} demonstrate superior performance at 512 resolution with a mean-F1 of $86.14$, which is $0.07$ higher than DML\_CSR. We achieve these results without training on multiple resolutions, i.e., we train on just $256\times 256$ and the network seamlessly scales to multiple resolutions. 
Our results on the Helen dataset, which has a small number of training samples (2000), are in Table \ref{table:helen_results}. Our performance is close to SOTA despite training on non-aligned face images.
Last but not least, in Table \ref{modelsize}, we comprehensively compare the model sizes, GFlops and FPS of all our baselines. With only 2.29 million parameters, \OURNAME is the most compact face-parsing network available; it is 65 times smaller than FARL and 26 times more compact than DML\_CSR. Our fps, evaluated on an Nvidia A100 GPU, stood at 110 frames per second, whereas DML CSR's performance was at 76 frames per second, and EAGR and AGR-Net demonstrated fps of 71 and 70, respectively. Additional comparative analysis of our results with DML\_CSR  is included in the supplementary.
 % Table \ref{tab:comparison_results} shows the results of different approaches 
% \noindent \textbf{Qualitative Visualizations.} 
% In this section, we discuss the key differences between the qualitative  results is illustrated in ~Fig.  \ref{fig:vis_forms_plus_pub}. 

\subsection{Ablations}
To evaluate the effect of several components we conduct the following ablations. 

\textbf{Network without LIIF Decoder: }We replaced the LIIF decoder with a Conv U-Net type decoder. The total parameter count of this model is 9.31M params (3x FP-LIIF). The Mean F1 for this model on LaPa dataset is 84.9 compared to 92.4 for our model. 

\textbf{EDSR ResBlock vs BN/IN ResBlocks:} The Old-EDSR ResBlock network produces an F1 of 92.3 on the LaPa dataset. New ResBlock + BN produces 92.32 and ResBlock + IN produces 92.4. The slight improvement prompted us to use IN.

\textbf{Edge-aware Cross-entropy and $\lambda$:} The table \ref{table:lambda} indicates the effect of $\lambda$ on the modulation of the edge-aware cross-entropy loss. 
\begin{table}[h]
\centering
\begin{tabular}{|l|l|l|l|l|l|}
\hline
$\lambda$      & 0     & 10   & 20    & 30    & 40   \\ \hline
F1 on LaPa & 91.73 & 92.2 & 92.29 & 92.34 & 92.4 \\ \hline
\end{tabular}
% \captionsetup{aboveskip=4pt,belowskip=-10pt}
\caption{Effect of edge-aware loss modulated by $\lambda$ on networks performance}
\label{table:lambda}
\end{table} 

\textbf{Comparison with lightweight segmentation model:}
Table \ref{table:sfnet} shows the results for face segmentation on LaPa using SFNet\cite{sfnet} which is a recent lightweight segmentation network for cityscapes.
\begin{figure}[!h]
    \centering
    \scalebox{0.9}{
    \begin{tabular}{|c|c|c|c|c|c|c|}
    \hline
    Class & SFNet & Ours && Class & SFNet & Ours \\
    \cline{1-3} \cline{5-7}
    Skin & 94.75 & 97.6 && R-Eye & 76.12 & 92.2 \\
    Hair & 87.27 & 96.0 && L-Brow & 76.98 & 90.90 \\
    Nose & 98.71 & 972. && R-Brow & 73.8 & 90.60 \\
    I-Mouth & 78.86 & 90.30 && U-Lip & 97.28 & 87.8 \\
    L-Eye & 79.55 & 92.00  && L-Lip & 96.23 & 89.5 \\
    \hline
    \multicolumn{5}{|r|}{Mean} & 85.96 & 92.41 \\
    \hline
    \end{tabular}
    }
    \caption{Comparison with SFNet \cite{sfnet}, another lightweight segmentation network.}
    \label{table:sfnet}
\end{figure}
\vspace{-6pt}
\subsection{Low Resource Inference}
\label{sec:lowresource}
One of the practical advantages of \OURNAME is that it enables face parsing on low-resource devices. The primary prerequisite to enable
low resource inference is that a model’s inference should be low in compute and therefore have a high frame per second (FPS)
count. Our ability to predict segmentation masks at multiple resolutions enables us to meet the demand for low inference costs. To achieve this, we can instruct the network to perform a low-resolution prediction, and the result can be upscaled to a higher resolution. Table
\ref{table:fps-table} shows the FPS for lower-resolution inference on a single Nvidia A100 GPU. However, the shorter inference time should not result in poor quality output when upsampled to a higher resolution. Therefore we compare our upscaled outputs with the ground truth and
present the findings in Table \ref{table:lapa_results}, \ref{table:celeba_results}. Here, we can see that our 192$\times$192 or 128$\times$128 segmentation output, when upscaled to 256$\times$256, leads
to a minimal loss in quality, as can be seen in both classwise and overall F1 scores. 
% Further, it can be seen that the resolution of 192$\times$192 provides the best trade-off between FPS and F1 score. 
In Table \ref{table:lapa_results}, \ref{table:celeba_results} Ours$^{192\rightarrow256}$ denote results of upsampling from 192$\times$192 to 256$\times$256 and similarly Ours$^{128\rightarrow256}$,Ours$^{96\rightarrow256}$ and Ours$^{64\rightarrow256}$ denote results of upsampling from 128, 96 and 64 respectively to 256$\times$256. In addition to faster inference, a low parameter count or smaller size model  helps in model transmission under low bandwidth circumstances.
\begin{table}[!ht]
\centering
\scalebox{0.9}{
\begin{tabular}{|c|c|c|c|c|c|}
\hline
Res & 64x64 & 96x96 & 128x128 & 192x192 & 256x256 \\ \hline
FPS        & 445   & 332   & 294     & 187     & 110     \\ \hline
FLOPS      & 27.44 & 32.25 & 39      & 58.24   & 85.2    \\ \hline
\end{tabular}
}
\caption{Frame Per Second for different resolution output while keeping the input image resolution constant at 256$\times$ 256}. 
\label{table:fps-table}
\end{table}
\vspace{-0.5cm}
\subsection{Limitation}
Encouraged by the performance of this low parameter \OURNAME network in face segmentation, we tested its effectiveness at semantic segmentation in a more generic domain like Cityscapes \cite{Cordts2016Cityscapes}. We chose this dataset because it lacked the structural regularity that we exploited in this work and segmentation using the current architecture should not be feasible. As expected, the mIoU score on the validation was reported at 62.2, which is 20+ points lower than SOTA models reporting scores in the range of $\sim$85 mIoU. 
% Another aspect of \OURNAME is the inference speed
% \todo{conv decoder} 
% \todo {no edge loss}
% \todo {multi resolution training}
% \label{sec:ablation}

\section{Conclusion and Future Work}
This work presents \OURNAME, an implicit neural representation-based face parsing network. We exploit the human face's regular and locally consistent structure to propose a low-parameter, local implicit image function-based network that can predict per-pixel labels for a human face image. Our model is $1/26^{th}$ or lesser in size compared to the current SOTA models of face parsing but outperforms them on mean-F1 and mean-IoU over multiple datasets like CelebAMask-HQ and LaPa. This network can also generate outputs at multiple resolutions, which can be very useful in reducing the inference time of segmentation by generating output at a lower resolution. These properties make it feasible to use this architecture on low-resource devices. 

Future work will address misprediction in regions with fewer class labels. We would also extend Implicit neural representation-based segmentation to domains with a regular and consistent structure, like medical imaging, human body parts, etc., and to domains where structure uniformity is not guaranteed, like in the wild images.

%-------------------------------------------------------------------------

%-------------------------------------------------------------------------

%%%%%%%%% REFERENCES
{\small
\bibliographystyle{ieee_fullname}
\bibliography{egbib}
}
\newpage
\pagebreak
\onecolumn
\twocolumn

\subsection{Analysis}
\begin{figure}[!ht]
\begin{center}
  \includegraphics[width=\linewidth]{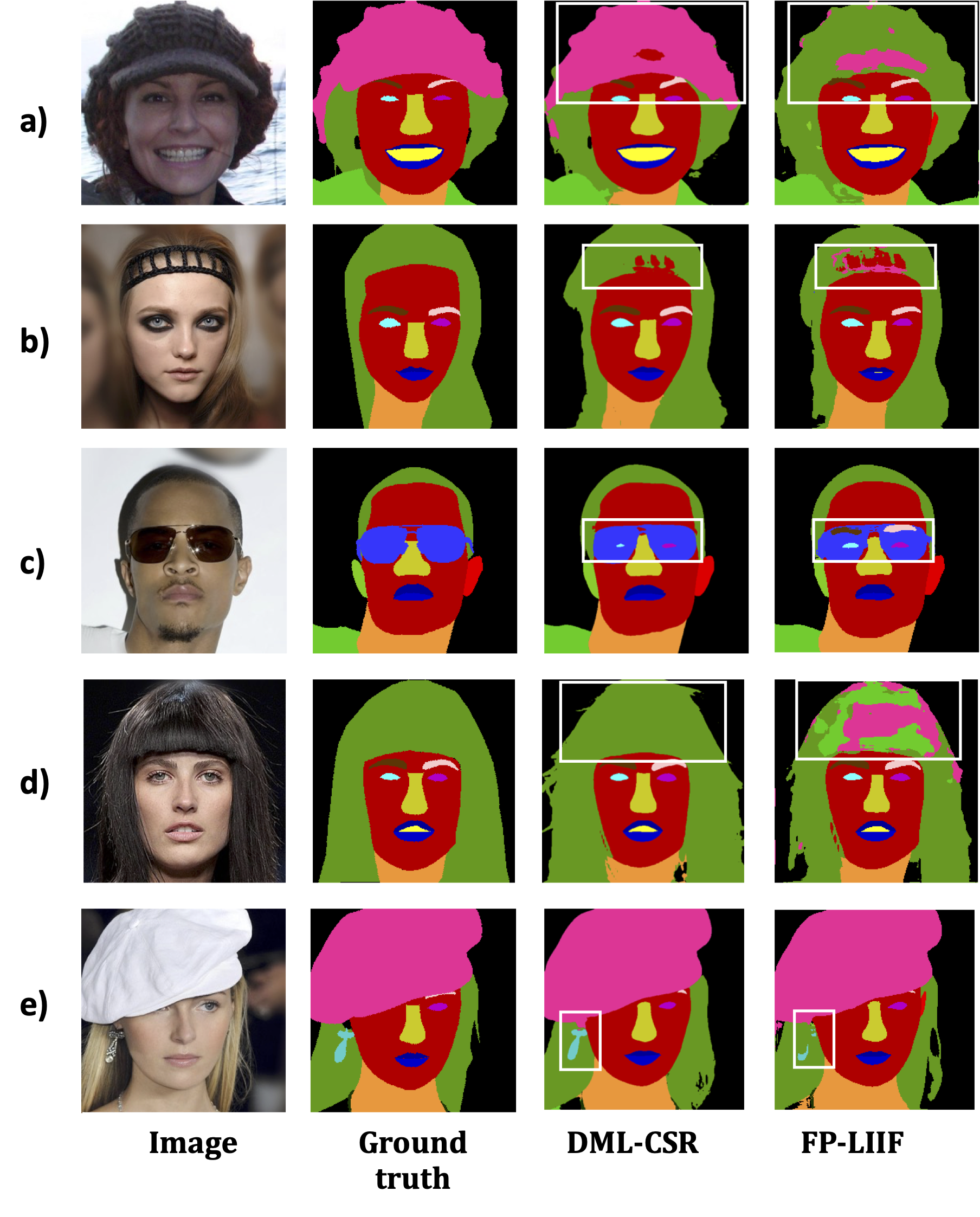}
\end{center}
\caption{Few results where DML\_CSR performed better than \OURNAME on CelebAMask-HQ dataset.} 
\label{fig:celeb-dml-better}
\end{figure}
\begin{figure}[!ht]
\begin{center}
  \includegraphics[width=\linewidth]{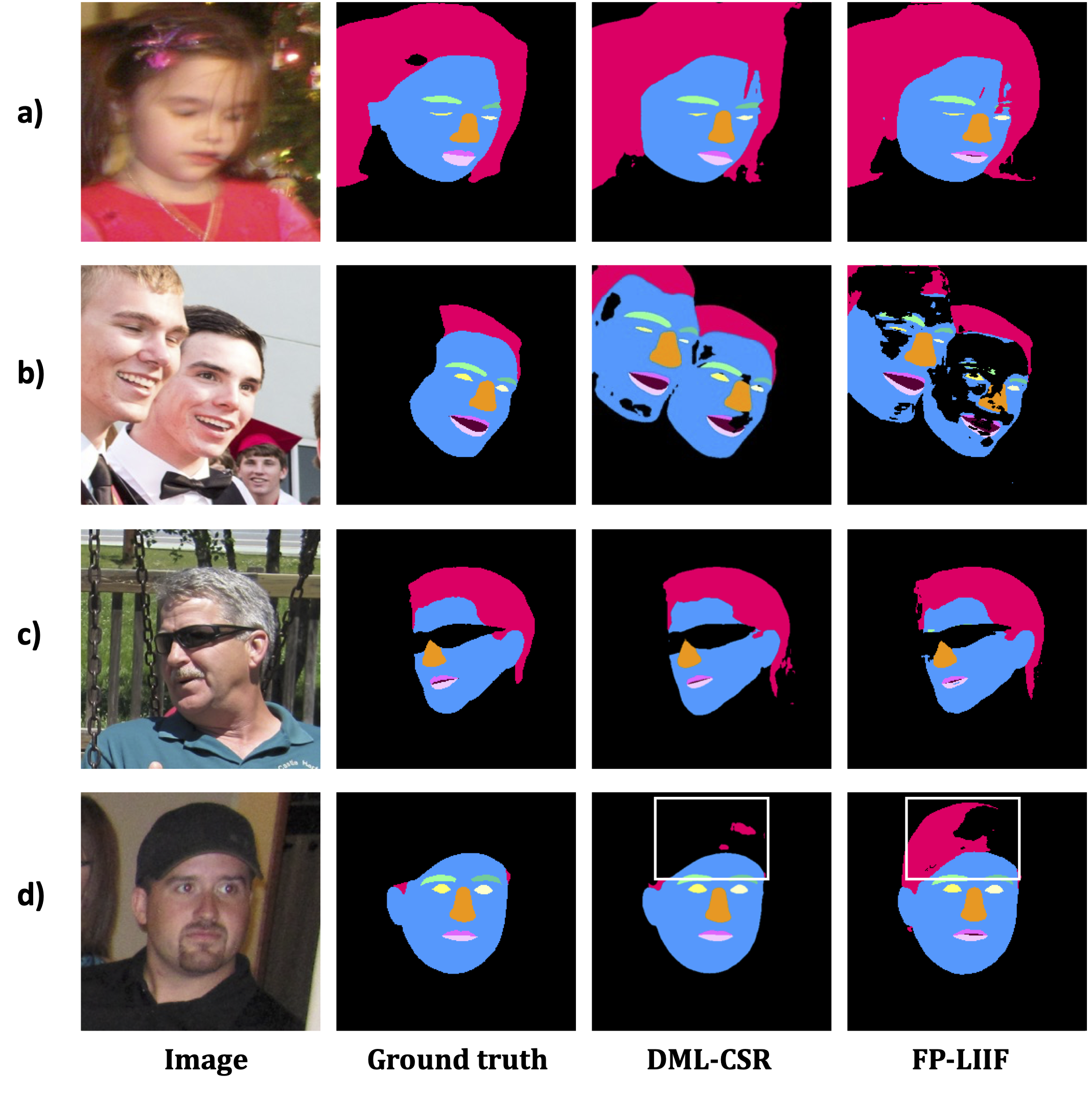}
\end{center}
\caption{Few results where DML\_CSR performed better than \OURNAME on LaPa dataset} 
\label{fig:lapa-dml-better}
\end{figure}

The quantitative results shown in Table \ref{table:lapa_results}, \ref{table:celeba_results} points that
even though \OURNAME fares better in mean F1, the best classwise performance is scattered across multiple models. But at the same time, the gap between the best classwise scores and \OURNAME's classwise scores is marginal. Therefore, we try to further identify the problematic areas and include visualizations of \OURNAME's worst-performing results compared to DML\_CSR in F1 in Figure \ref{fig:lapa-dml-better}, \ref{fig:celeb-dml-better}. It can be seen from Figure \ref{fig:lapa-dml-better} that rows a) and c) have negligible differences, and in the remaining rows, both are performing poorly in the problematic regions of hair and face. In Figure \ref{fig:celeb-dml-better}'s rows b) and d), the F1 scores for these are debatable because of incorrect labeling in the ground truth. In the remaining rows, the underrepresented class of hat and earrings are bringing down our performance. Therefore the current setup of \OURNAME is affected by a lack of data as compared to DML\_CSR. This can also be corroborated by Table \ref{table:celeba_results}. It is also necessary to point out that the ground truth data of CelebAMask-HQ is noisy (Figure \ref{fig:gt-worse}) and can cause problems in training and testing.

From the point of view of inference time, \OURNAME could be used to generate segmentation at a lower resolution, and the generated output scaled at the required higher resolution to improve inference time and hence increase fps. The generation of lower-resolution segmentation does not require any additional training and is an outcome of being an implicit neural representation network. The 128-resolution version of \OURNAME clocked an fps of 294 compared to the regular version of resolution 256, which runs at 120 fps. This makes our model more conducive for low compute devices. 

\subsubsection{Variance in performance over mulplte runs}
We also calculate the mean and variance of our model's F1 score for Lapa, CelebAMask-HQ and Helen in Table \ref{table:mean_var}
\begin{table}[!ht]
\centering
\begin{tabular}{|l|l|l|}
\hline
        & Mean        & SD          \\ \hline
F1 Lapa & 92.35 & 0.06 \\ \hline
F1 Celeb & 85.90 & 0.20 \\ \hline
F1 Helen & 91.12 & 0.10 \\ \hline
\end{tabular}
\captionsetup{aboveskip=4pt,belowskip=-10pt}
\caption{Mean and Variance of \OURNAME}
\label{table:mean_var}
\end{table}
 It should be noted that other state-of-the-art works do not report these mean and variance over multiple runs and therefore direct comparison of these numbers is not possible.
\end{document}